# Bilateral Deep Reinforcement Learning Approach for Better-than-human Car-following

Tianyu Shi[1*], Yifei Ai[2*], Omar ElSamadisy[1,3*], Baher Abdulhai[1]

*Abstract*—Car-following based on Reinforcement Learning (RL) has received attention in recent years with the goal of learning and achieving performance levels comparable to humans, based on human car following data. However, most existing RL methods model car-following as a unilateral problem, sensing only the leading vehicle ahead.

For better car following performance, we propose two extensions: (1) We optimise car following for maximum efficiency, safety and comfort using Deep Reinforcement Learning (DRL), and (2) we integrate bilateral information from the vehicles in front and behind the subject vehicle into both state and reward function, inspired by the Bilateral Control Model (BCM). Furthermore, we use a decentralized multi-agent RL framework to generate the corresponding control action for each agent. Our simulation results in both closed loop and perturbation tests demonstrate that our learned policy is better than the human driving policy in terms of (a) inter-vehicle headways, (b) average speed, (c) jerk, (d) Time to Collision (TTC) and (e) string stability.

## I. INTRODUCTION

Inherent system randomness in human-driving behavior [1] creates instability in the traffic system. Shockwaves and stop-and-go have become a primary safety concern and the main cause of traffic jams [2]. Meanwhile, human drivers also want to maximize travel efficiency, such as improving average speed and minimizing headway [3]. As a result, a critical question for building an intelligent car following system is how to encourage the vehicle to travel as fast as possible while maintaining safe efficient headway to the leading vehicle and reducing shockwaves.

Autonomous driving technology has been studied for years and started to come to reality with the development of sensors and Artificial Intelligence (AI). The autonomous driving vehicle could potentially learn to outperform human driving in safety and comfort [3][5]. One major benefit of Connected and Autonomous Vehicle (CAV) is that the randomness in driving behavior can be significantly reduced; thus, the whole system can be better managed by control algorithms with minimum reaction time.

Car-following is a critical driving task. Many models have been developed to mimic human driving behavior [7][8]. In traffic flow theory, classic Car-Following Models (CFMs) are based on physical knowledge and human behaviors, etc.

For example, Gipps model considers both free-flow mode (without leading vehicle) and car-following mode (with the leading vehicle) and takes the minimum velocity of them to decide whether to apply acceleration or deceleration. The following vehicle's speed is also limited by safety constraints [7]. Another well-known model is the Intelligent Driver Model (IDM), which models the output acceleration based on the desired velocity, headway, relative velocity, and distance to the leading vehicle [8].

In recent years, some studies proposed data driven method to train CFMs. He et al. [9] used K-Nearest Neighbors (KNN) to find the most likely behavior of vehicles in the car-following mode. Some studies also apply supervised learning. Chong et al.[10] used Neural Networks (NN) to model driver behavior regarding to longitudinal and lateral actions. Zhou et al. [11] focused on capturing and predicting traffic oscillation using Recurrent Neural Networks (RNN) [12].

Although supervised learning methods have shown very good results, it requires hand-collected microscopic car-following data which are rare and expensive to collect. Usually, collected data is from human drivers. Some learning methodologies such as Imitation Learning might also lead the trained models to learn some irrational behavior, such as very aggressive or very conservative behaviors from humans. Applications of RL have rapidly matured in recent research [21]. RL has successfully addressed problems such as Go [13] and Atari games [14]. In such framework, RL agents interact with the environment and observe the state and the corresponding reward. They are expected to find the optimal policies that maximize accumulated reward after training. A well-defined reward function usually makes agent's policy converge to the objective we want to achieve. Zhu et al. [15] defined CFMs to keep safe distance to the front vehicle and maintain desired speed. Their proposed model can achieve good generalization ability compared to data-driven CFM. However, how to develop better than human driven CFM [8] is still needed for further investigation.

In this paper we focus on 1 and 2. We address CFM to meets efficiency, safety, comfort, and traffic stability performance goals.

Previous CFMs only consider the information from the leading vehicle [5][7][8][15]. Wang and Horn [16] first proposed the BCM into traffic flow control by adding information about the following vehicle. In their results, they demonstrate better string stability under the BCM compared to

[1] Tianyu Shi, Omar ElSamadisy, Baher Abdulhai are with the Department of Civil & Mineral Engineering, University of Toronto, Toronto, Ontario, Canada. ty.shi@mail.utoronto.ca, omar.elsamadisy@mail.utoronto.ca, baher.abdulhai@utoronto.ca
[2] Yifei Ai is with the Department of Mechanical & Industrial Engineering, University of Toronto, Toronto, Ontario, Canada. yifei.ai@mail.utoronto.ca

[3] Omar ElSamadisy is on leave with Department of Electronics & Communications Engineering, College of Engineering and Technology, Arab Academy for Science, Technology, and Maritime Transport, Alexandria, Egypt. omar.elsamadisy@aast.edu

* Indicates equal contribution

pure CFM traffic. In terms of chain stability, BCM demonstrates better performance than CFM. However, it's still unknown whether the BCM model will demonstrate better travel efficiency, comfort, and safety than CFM. More specifically, BCM model can't learn how to balance and optimize efficiency, safety, comfort, and platoon stability together.

In this work, we extend the BCM concept into a DRL framework to solve the multi-objective problem. Our experimental results demonstrate that our proposed bidirectional DRL CFM can surpass previous human driving models in terms of (1) inter-vehicle headways, (2) average speed, (3) jerk, (4) safety and (5) string stability.

## II. PROBLEM FORMULATION

### A. Evaluation criteria

To evaluate the performance of our CFM, we illustrate the following metrics.

*1) TTC:* Safety is one of the most important features in driving systems. We use TTC to represent safety. TTC represents the time left before two vehicles collide. It is computed as [5]:

$$TTC(t) = \frac{S_{n-1,n}(t)}{\Delta V_{n-1,n}(t)} \quad (4)$$

where $t$ stands for time; $n-1$ is the index of the leading vehicle, $n$ is the index of the following vehicle; $S_{n-1,n}(t)$ denotes the clearance distance between the leading and the following vehicles; $\Delta V_{n-1,n}(t)$ represents the relative speed at time t: $V_{n-1}(t) - V_n(t)$.

*2) Headway:* Efficiency of a CFM can be measured by resulting average headway, the less the average headway the more the resulting road capacity. Time headway is defined as the time it takes for the following vehicle to reach the current location of the leading vehicle. Let $H(t)$ denote the headway $h$ at time $t$, then:

$$H(t) = \frac{S_{n-1,n(t)} + l_{n-1}}{v_n(t)} \quad (5)$$

where $l_{n-1}$ is the length of the leading vehicle. $S_{n-1,n(t)}$ is the clearance distance, $v_n(t)$ is the velocity of the following vehicle. In the experiment, we will use average headway over one episode as the efficiency metric

*3) Jerk:* Comfort is another important element in autonomous driving. We use the jerk to measure comfort. Jerk is defined as:

$$Jerk(t) = \frac{a_n(t) - a_n(t-1)}{\Delta t} \quad (6)$$

where $\Delta t$ is the time interval between time steps $t$ and $t-1$, $a_n(t)$ is the acceleration of vehicle $n$ at time $t$. In the experiment, we use average absolute jerk over one episode as the metric.

### B. CFMs

We will first illustrate some classical CFMs, including both unilateral and bilateral CFMs to compare with our bilateral RL CFM.

Given the leading vehicle (L), the current vehicle (C) and follower vehicle (F), A CFM controls the longitudinal movements for the current vehicle with respect to the leading vehicle's motions.

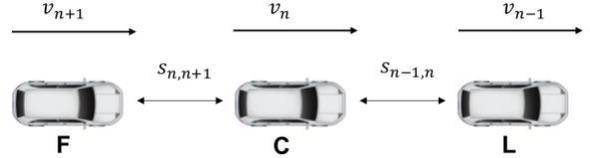

Figure 1. Illustration of the difference between CFM and BCM. 'L' represents the leading vehicle, 'C' represents the current vehicle, 'F'represents the following vehicle [5].

Several CFMs have been proposed to model the microscopic traffic behavior. As shown in Figure 1. if it is unilateral model, it will only focus on the information from the leader ahead; if it is bilateral model, it will utilize both the information from leader ahead and follower behind.

*1) Gipps [1] (unilateral model):* Gipps is a classical model for modeling driving behavior.

Gipps considers both free-flow mode (without leading vehicle) and car-following mode (with the leading vehicle) and takes the minimum velocity of them.

$$v(t + \Delta t) = min\ [v + a\Delta t, v_0, v_{safe}(s, v_l)] \quad (7)$$

where $v_0$ is the desired speed, $\Delta t$ is the reaction time, and $v_{safe}$ is the safe speed, for which if the leading vehicle decelerates to a complete stop, the $v_{safe}$ of the following vehicle needs to guarantee not to collide with leading vehicle.

*2) IDM [8] (unilateral model):* IDM models realistic acceleration profiles in the single-lane traffic situations. The following acceleration equation is given below:

$$\dot{v}_c = a\left[1 - \left(\frac{v_c}{v_0}\right)^\delta - \left(\frac{s^*(v_c, \Delta v_c)}{s}\right)^2\right] \quad (8)$$

where $v_0$ is the desired velocity for the current vehicle, $v_c$ is the current velocity for the current vehicle, $s$ is the current distance to the leading vehicle, $\delta$ is usually set as 4) [8]. $a$ is the maximum vehicle acceleration, where $-a\left(\frac{s^*(v_c,\Delta v_c)}{s}\right)^2$ is a breaking term, in which the desired distance $s^*$ is defined as,

$$s^*(v_c, \Delta v_c) = s_0 + max\left(v_c T + \frac{v_c \Delta v_c}{2\sqrt{ab}}, 0\right) \quad (9)$$

where $s_0$ is minimum jam distance, which is set as 2m, $T$ is the desired time headway, and $a$ is the maximum vehicle's acceleration, $b$ is comfortable deceleration.

*3) BCM [16] (bilateral model):* BCM assumes the vehicle has an additional back sensor. Its control is based on the state

of both the leading vehicle and the following vehicle as shown in Figure 1. The controller's equation is given below:

$$a_n = k_d(s_{n-1} - s_n) + k_v(r_{n-1} - r_n) \quad (10)$$

where the $s_{n-1}$ denotes the space between the current vehicle and its leading vehicle; $s_n$ denotes the space between the current vehicle and its follower vehicle. $r_{n-1}$ and $r_n$ denotes the relative speed between the current vehicle to its leading vehicle and its follower vehicle respectively. The $k_d$ and $k_v$ are proportional and derivate gains respectively. Note that when there is no following vehicle, it will switch to unilateral car-following mode:

$$a_n = k_d(s_{n-1} - s_0) + k_v(v_{n-1} - v_n) \quad (11)$$

where we want to control the relative space and speed between the current vehicle to its leader. $s_0$ is the desired space that can be set adaptively according to the car's speed, i.e., $s_0 = v_n T$ and $T$ is the reaction time. And $v_{n-1}$ is the speed of current vehicle (Vehicle C in Figure 1) and $v_n$ is the speed of leader vehicle (Vehicle L in Figure 1. This formula will be the same as shown in the unilateral CFM [24] which is similar to Shladover model [24].

## III. METHOD

### A. Bilateral DRL for car-following

In the car-following scenario, if the lead vehicle speed oscillates, the oscillation can potentially magnify in the platoon of following vehicles leading to unstable platoon behavior. Therefore, one of our four objectives is to make sure such oscillations are damped as soon as possible in the following platoon. We can define our problem following the BCM logic as suggested by [16]. According to [16], BCM chains (a chain of vehicles controlled by BCM) can make the perturbation from leading vehicles decay exponentially. Therefore, we augment our state space using bilateral sensing information which is described next.

### B. State and Action Design

*1) State Space:* Based on [5], the RL-based CFM designs the state $s_t$ as a 5-dimensional array: $(v_n(t), S_{n-1,n}(t), \Delta V_{n-1,n}(t), v_l, a_n(t-1))$ where $v_n(t)$ is the following vehicle speed at time step $t$, $S_{n-1,n}(t)$ is the clearance distance, $\Delta V_{n-1,n}(t)$ is the relative speed of vehicle $n-1$ with respect to vehicle $n$, $v_l$ is the target speed on the current section of the road, and $a_n(t-1)$ is the acceleration of the following vehicle $n$ from last step. Unlike the CFM-based RL model [5], we add two additional dimensions into our state space: clearance distance of the current vehicle $n$ to its following vehicle $n+1$: $S_{n,n+1}(t)$; the relative speed of vehicle $n$ with respect to vehicle $n+1$: $\Delta V_{n,n+1}(t)$. Therefore, our state space $s_t$ is a 7-dimensional array:

$$s_t = \left(v_n(t), S_{n-1,n}(t), \Delta V_{n-1,n}(t), S_{n,n+1}(t), \Delta V_{n,n+1}(t), v_t, a_n(t-1)\right) \quad (12)$$

*2) Action Space:* Action space $A$ is a continuous 1-dimensional space with range $(-3m/s^2, 3m/s^2)$, it represents the acceleration of the controlled RL vehicle.

### C. Reward Function Design

We design the reward function based on safety, efficiency, and comfort. We evaluate the agents based on these three rewards as well as platoon stability.

*1) Safety $f_{safety}$:* We want to penalize vehicles with small TTC. At the same time, a vehicle with reasonable TTC should not be penalized. Thus, it is important to determine a threshold. According to [5], we use the following function as the safety component in the reward function:

$$f_{safety}(ttc) = \begin{cases} \log\left(\frac{ttc}{4}\right) & 0 \le ttc \le 4 \\ 0 & othwewise \end{cases} \quad (13)$$

*2) Efficiency $(f_{eff})$:* For the efficiency component, we use a log-normal distribution on based on the time headway defined in equation 5. Let $H \sim Log-normal(u, \sigma)$, then we have the following Probability density function (pdf) for $H$:

$$f_{eff}(h) = \frac{1}{\sqrt{2\pi} h \sigma} \exp\left(-\frac{(\ln(h) - u)^2}{2\sigma^2}\right) \quad (14)$$

According to [5], based on the empirical data, we choose $u = 0.4226$, $\sigma = 0.4365$. As shown in [5], the maximum is located at some point around $h = 1.26$ with maximum value around 0.659. Later in the paper we use lower values of h to achieve better than human performance, but in this section we use the same value as in [5] for fair comparison.

*3) Comfort $(f_{comfort})$:* Comfort component is based on jerk, which is the rate of change of acceleration:

$$f_{comfort}(jerk) = -\frac{jerk^2}{3600} \quad (15)$$

We want the range of this reward component to be similar to others. In our simulation, the seconds per simulation step is $0.1s$, thus the range of jerk is $(-60, 60)$. By dividing 3600, we can scale the range of $f(jerk)$ to $(-1, 0)$.

Therefore, our reward function is defined as the sum of these reward components which is used in CFM-based RL model:

$$R_{CFM} = \omega_s * f_{safety} + \omega_e * f_{eff} + \omega_c * f_{comfort} \quad (16)$$

where $\omega_s, \omega_e, \omega_c$ are the weights for safety, efficiency, comfort functions. We set $\omega_s = \omega_e = \omega_c = 1$.

*4) Following vehicle reward objective:* Besides those three reward components in the front view CFM, we would also add two additional back-view reward components:

- **Following Vehicle Safety $(f_{safety-f})$**: Measure the safety of the following vehicle $n+1$ based on the TTC of the current vehicle $n$ and the following vehicle $n+1$. The function is the same as the safety reward function (Equation 13) in the CFM.

- **Following Vehicle Efficiency $(f_{eff-f})$**: Measure the efficiency of the following vehicle $n+1$ based on the time headway of the following vehicle $n+1$. The function is the same as the efficiency reward function (Equation 14) in the CFM.

Therefore, our final reward function will be:

$$r = \omega_s * (f_{safety} + f_{safety-f}) + \omega_e * (f_{eff} + f_{eff-f})$$
$$+ \omega_c * f_{comfort} \quad (17)$$

For a fair comparison, the weights are set as the same as in CFM model.

### D. Learning Scheme

The learning agent interacts with the environment, which runs on the SUMO [20] simulator. We also use the FLOW [22] Python library to interact with the SUMO simulator. In the experiment, we consider all human driving vehicles are controlled by IDM model as same in FLOW[22]. We run 120 episodes and each episode has 3600 steps.

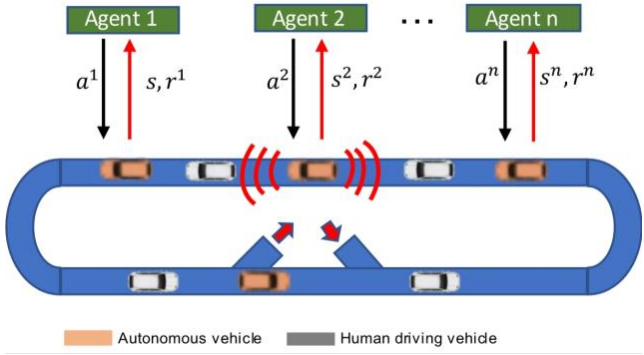

Figure 2. Distributed learning framework for bilateral RL CFM. The road network is a single-lane closed-loop network with an entrance and exit ramp. During the training, the red vehicle will be driving on this closed-loop road. The gray vehicles will be generated by the simulator and flow into the network from the bottom-right ramp and leave the road by driving to the bottom-left ramp. The inflow rate of the human driving vehicle is 180 veh/h, and there are 5 autonomous vehicles in the network. The total vehicle number is 10.

Training in a multi-agent scenario is challenging. Firstly, the state and action space will grow exponentially with the increase of autonomous vehicle agents by using a CL training approach [22]. Secondly, we assume that there is no V2V or V2I communication so that each autonomous vehicle agent will only have a local observation.

We select the training network as the closed-loop network, which is shown in Figure 2. For the loop network, it's easy to add inflow and outflow, which is helpful for stabilizing the learning. Secondly, the agent can gather more experience in a loop network compared to an open road network because the agent won't run out of the network.

Our DL multi-agent training framework is shown in Figure 2. There are $n$ agents in the simulation scenario. Each agent is controlled in a distributed framework, i.e., each agent will only sense the leading and following vehicle, then get the corresponding observation $s^i$ and the reward $r^i$. The policy is shared among all agents in the traffic network. The learned policy will generate the control action $a^i$ for each agent.

## IV. EXPERIMENTS

### A. Baseline selection

In the experiment, we consider the following methods as shown in Table I. Firstly, we consider a CL multi-agent version of the BCM-RL method, named BCM-CLRL in comparison with our method (BCM-DLRL) to compare different learning frameworks. Secondly, we consider another car-following RL baseline from [5]. Then we consider several non-learning car-following baselines which include BCM [16], Gipps [7], and IDM [23].

TABLE I.

Summary for different baseline methods

| methods | learn | safe | Efficient | comfort | Stable |
|---|---|---|---|---|---|
| CFM-CLRL [5] | CL | Yes | Yes | Yes | No |
| BCM [16] | No | No | No | No | Yes |
| Gipps [7] | No | No | Yes | No | No |
| IDM [8] | No | No | Yes | Yes | No |
| BCM-CLRL (ours) | CL | Yes | Yes | Yes | Yes |
| BCM-DCRL (ours) | DL | **Yes** | **Yes** | **Yes** | **Yes** |

Firstly, we test the trained agent in the same closed loop road network. In the training environment, there are ten vehicles in total and five of them are autonomous vehicle agents. We keep the same agent numbers for both training and testing scenarios. These analyses are given below.

*1) Safety analysis:* Safety is an essential element for evaluating performance. TTC is one surrogate indicator to represent safety. To compare the performance, we plot the TTC distribution for each method.

By comparing Figure 3, we find that BCM's safety is better than the other two car-following baselines. BCM-CLRL is slightly better than CFM-CLRL. Combined with a DL multi-agent learning framework, the performance further improves.

*2) Efficiency analysis*: To compare the efficiency for different methods, we calculate the headway at each time step. At later time steps the vehicle platoon becomes more stable and headways become smaller.

From Figure 4, we can find that BCM model tends to have unstable headway. Our method (BCM DCRL) shows the lowest headways and hence most flow-efficient car-following.

*3) Comfort analysis:* To visualize the jerk under a control policy, we plot the evolution of acceleration for each method given one episode.

From Figure 5. , we can find that BCM and Gipps tend to have abrupt acceleration and hence higher jerk. We conjecture it's because these two models do not consider action bounds.

*4) Summary:* To make a comprehensive comparison, we select the average headway(s), jerk(m/s2), TTC(s) over one episode to measure the efficiency, comfort, and safety.

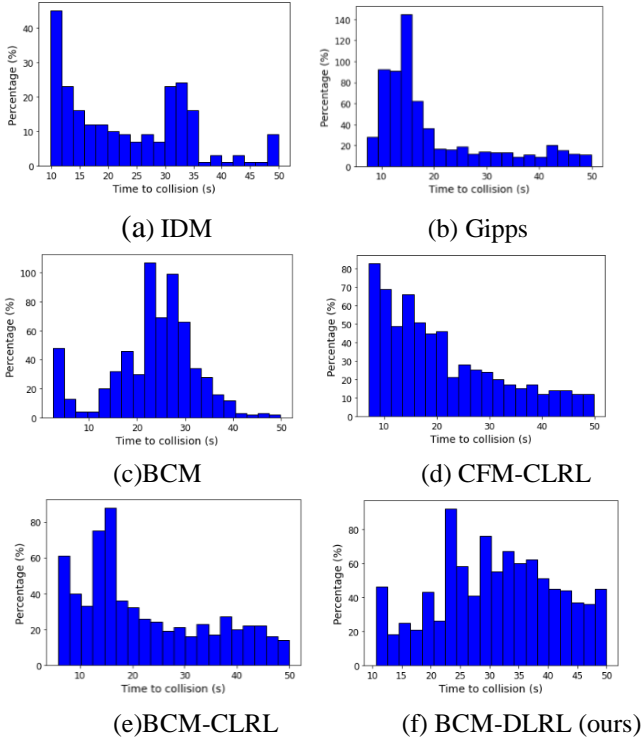

(a) IDM   (b) Gipps
(c) BCM   (d) CFM-CLRL
(e) BCM-CLRL   (f) BCM-DLRL (ours)

Figure 3.  Distribution of TTC during closed loop car-following. The target headway for each baseline is controlled as 1.26s. The more close to 0, the more unsafe.

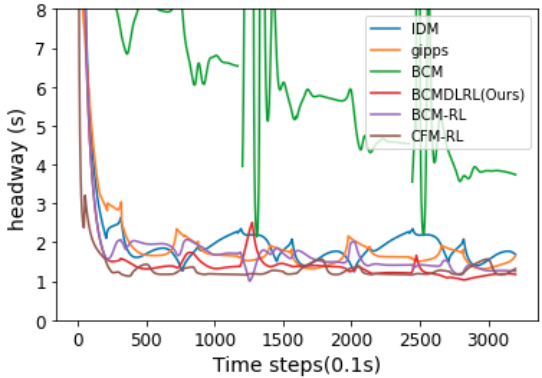

Figure 4.  Headway comparison in closed loop car-following scenario. The target headway is 1.26s.

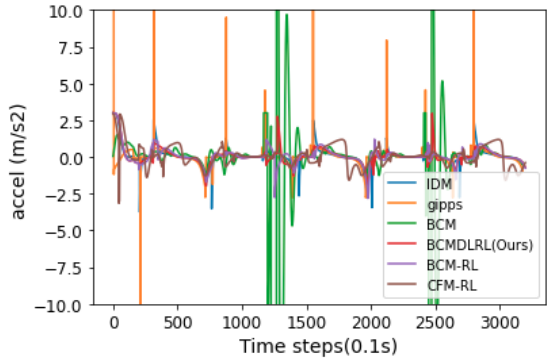

Figure 5.  Accleration comparison in closed loop car-following scenario.

From TABLE II. , among the non-learning baselines, we find that the Gipps model is more aggressive than IDM and BCM, as it has the smallest headway but the highest jerk. For learning-based models, comparing BCM-CLRL and CFM-CLRL, we find that BCM-CLRL sacrifices efficiency, i.e., uses higher headway, to achieve higher comfort and safety. As a result, they will have a smaller jerk and higher TTC. We also find BCM-DLRL method and CFM-CLRL are close to the target headway, i.e., 1.26 s, but the safety and comfort of our model are better than CFM-CLRL. In the next section we assess platoon stability across the models.

TABLE II.

Comparison of different methods in closed loop test

| Methods | Headway | Jerk | TTC |
| --- | --- | --- | --- |
| CFM-CLRL[5] | **1.259** | 0.215 | 21.037 |
| BCM [16] | 2.175 | 0.341 | 28.970 |
| Gipps [7] | 1.755 | 1.400 | 19.714 |
| IDM [8] | 2.669 | 0.229 | 23.181 |
| BCM-CLRL (ours) | 1.732 | 0.173 | 22.372 |
| BCM-DLRL (ours) | **1.285** | **0.163** | **31.176** |

### B. Perturbation test experiment

To complete the comparisons, we subject a platoon of vehicles on a straight road segment to traffic perturbation and visualize the platoon stability across all models. The focus of BCM, as a concept, is the suppression of the perturbation in the traffic. We visualize the behavior of the platoon using the space-time diagram shown in Figure 6. To create perturbations, we oscillate the speed of the leading vehicle as shown by the red profile. The follower vehicles are plotted in blue. We compare the cases of the follower vehicles in the platoon being controlled by different CFMs as shown in Figures 6 a-f.

To make a comprehensive comparison, we also report the average headway, jerk, TTC over one episode to quantify the efficiency, comfort, and safety for each method. Note that because TTC values are very large due to the minor speed variations in this experiment, we compare safety using the log of TTC as in Equation 13.

Based on Figure 6, we observe two effects: how quickly the oscillations dampen and what is the resulting performance in terms of headway (is the platoon compact or spaced up), as well as jerk and TTC. Ideally, a better model would yield faster dampening of oscillations and tighter platoons with shorter headways. Further, lower jerk and higher log TTC are better. Visually from Figure 6 and numerically from Table III, we find that BCM models in general naturally dampen perturbations better. RL models in general achieve more compact platoons and shorter headways than non-learning methods. BCM-DLRL outperform CFM-CLRL because CFM-CLRL does not account for the effect of the follower vehicle, and hence exhibit less chain stability, which also reduces efficiency as well. BCM-DLRL further enhances chain stability over BCM-CLRL indicating better learning potential. We can find that our multi-agent BCM-DLRL model is better than CL-BCM-RL in terms of headway, jerk, and safety.

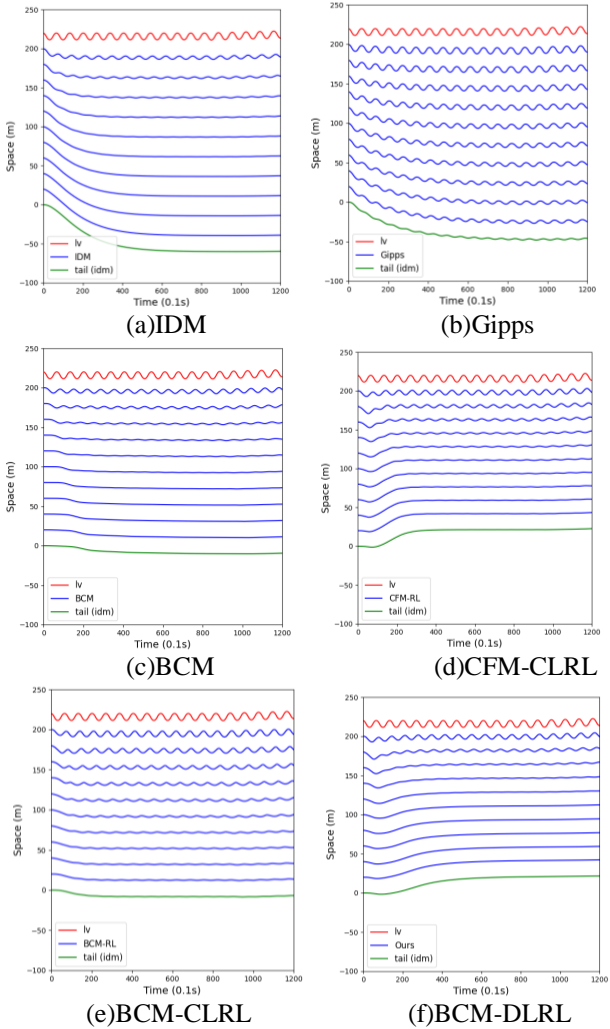

Figure 6. Perturbation of different controllers, larger oscillation means larger perturbation. The following vehicles move at speed of 20m/s, the number of BCM vehicle is 10. The y axis is the space which represent the relative position corresponding to given time steps. The leading vehicle and tail vehicle are controlled by IDM, other vehicles are controlled by the controllers in TABLE III. The tail vehicle is controlled by IDM because there is no vehicle behind to use BCM. For a fair comparison, we select IDM as the tail vehicle for all methods.

TABLE III.

Summary for perturbation test (target headway=1.26s)

| Methods | Headway | Jerk | Safety |
|---|---|---|---|
| CFM-CLRL [5] | 1.340 | 0.352 | 0.586 |
| BCM [16] | 1.361 | 0.300 | 0.422 |
| Unilateral-CFM[16] | 0.994 | 0.289 | 0.579 |
| Gipps [7] | 1.587 | 1.610 | 0.560 |
| IDM [8] | 1.675 | 0.236 | 0.491 |
| BCM-CLRL (ours) | 1.368 | 0.650 | 0.668 |
| BCM-DLRL (ours) | **1.180** | **0.057** | **0.827** |

TABLE IV.

Summary for perturbation test (target headway=0.8s)

| Methods | Headway | Jerk | Safety |
|---|---|---|---|
| CFM-CLRL [5] | 0.992 | 0.278 | -1.661 |
| BCM [16] | 1.361 | 0.300 | 0.422 |
| Unilateral-CFM[24] | 0.994 | 0.289 | 0.579 |
| Gipps [7] | 1.587 | 1.610 | 0.560 |
| IDM [8] | 1.158 | 0.368 | -1.301 |
| BCM-CLRL (ours) | 1.012 | 0.392 | 0.491 |
| BCM-DLRL (ours) | **0.830** | **0.345** | **0.581** |

Secondly, in Figure 6. we decrease the target headway to 0.8. A shorter target headway means a better efficiency. According to test data collected in the shanghai naturalistic driving study [5], the average value of human driving headway is 1.26. And in our experiment, our RL agents are encouraged to keep a target headway as 0.8, which is below the mode headway of human driving [5]. Note that the Shladover, Gipps, BCM models' formulas are independent on target headway so that they keep the same results. We observe CFM-CLRL[5] can achieve smaller headway than humans, but at the expense of lower safety. On the other hand, BCM-DLRL can achieve smaller headway than human driving while achieving better safety and comfort, demonstrating that it outperforms human driving performance. Finally, it is worth noting that for bilateral deep RL models, their average headway to the front vehicles is slightly worse than CFM whose average headway stays exactly at the optimal value. We believe this is a trade off since the bilateral RL models also need to maintain a good headway of its following vehicle.

Overall, our model outperforms all the other baselines in the perturbation test. While CFM achieves shortest headways, it does so at the expense of higher jerk and lower safety compared to BCM-DLRL.

V. CONCLUSIONS AND DISCUSSION

In this paper, we designed the bilateral DRL framework for car-following control. We find that our framework has better performance than human driving models. It is the most effective perturbation damper among these models. Also, it is in the top place in other metrics, i.e., safety, efficiency, comfort.

Notice that we also compare our DL multi-agent training framework with CL training of both CFM and BCM DDPG. We can also find that multiagent learning setting can further boost the performance.

In future work, we will add following target speed into the objective function. However, there would be challenges because more complicated multiple objectives will create learning challenges. Also, we will investigate how to make the autonomous vehicle agent become more adaptable to dynamic speed limit, i.e., If it receives a command to follow certain speed it will do even if it were in car following mode and even if the leading vehicle violates the set speed limit. Finally, we will test different penetration rate of autonomous vehicles to test the scalability of our method.